# Spatial Domain Feature Extraction Methods for Unconstrained Handwritten Malayalam Character Recognition

Jomy John[1]

[1] P M Government College, Chalakudy, Kerala, India
jomyeldos@gmail.com; jomy@pmgc.ac.in

**Abstract.** Handwritten character recognition is an active research challenge, especially for Indian scripts. This paper deals with handwritten Malayalam, with a complete set of basic characters, vowel and consonant signs and compound characters that may be present in the script. Spatial domain features suitable for recognition are chosen in this work. For classification, k-NN, SVM and ELM are employed

## 1 Introduction

Machine recognition of handwriting is challenging because of wide intrapersonal and inter-personal variation in human handwriting. In this paper we consider offline recognition of handwritten characters in Malayalam, a major Indian script. Studies on the recognition of handwritten Malayalam started quite recently [1] and most of the existing studies are limited to just basic characters [2]. The recognition task on this set becomes more challenging if there are a large number of classes with high similarity between some of them [3-5]. In this work, we have used spatial domain features such as topological features, distribution features, transition count features, local binary pattern descriptors and their combination as feature sets for the recognition of 90 character classes. Principal component analysis (PCA) is employed to reduce the dimension while combining the features. For classification, the classifiers such as k-Nearest Neighbour (k-NN), Support Vector Machines (SVM) with Radial Basis Function (RBF) and Polynomial kernel and two variants of Extreme Learning Machine (ELM) are used. The performances of various feature sets are analyzed for the large class classification problem.

The paper is organized as follows. Section 2 briefly introduces Malayalam language and script while section 3 and 4 explains different spatial domain features and classifiers used in this work. Experimental results and discussions are provided in section 5. Finally, section 6 concludes the paper.

## 2 Malayalam Language and Script

Malayalam is one of the twenty two scheduled languages of India. Spoken by 33 million people, it is the official language of the Kerala province, union territories of Lakshadweep and Puducherry. Its alphabet consists of vowels and consonants called basic characters. Each vowel has two forms, an independent form and a dependent form. There are 36 consonants and three special consonant signs. Apart from basic characters, there are pure and compound characters. Pure consonants are a unique property of Malayalam language and there are separate symbols to represent these pure consonants. Compound characters are special types of characters formed as a combination of two or more (pure) consonants. Two types of compound characters occur in Malayalam: vertically compounded and horizontally compounded. Though the script has been reformed, a mixture of old script and new script is noted in handwriting. Based on the above, we present in Fig. 1 the graphical shapes in the script namely independent vowels, dependent vowel signs, consonants, pure consonants, consonant signs and the entire compound characters in the modern script [6] along with the frequently used symbols from the old script, which are the subjects of this paper.

Fig. 1. Malayalam script character set

## 3 Methodology

A character recognition system [7] involves tasks such as pre-processing, feature extraction and classification. As pre-processing step, for each input image, we first find its bounding box and then normalize it to a square image of size 64×64 using bi-cubic interpolation technique. The gray scale values of the image are normalized with zero mean followed a Gaussian filter of size 3×3 to smooth the image. This image is then subject to feature extraction, which is explained in section 3.1 to 3.4.

### 3.1 Topological Features

**3.1.1 Number of loops**:

A loop is a handwritten pattern, when the writing instrument returns to a previous location giving a closed outline with a hole in the centre. There are many characters in Malayalam without any loop, with one loop, with two loops and so on. For example, in the case of the character pattern ര (/ra/), ണ (ṇa), ഞ (ña) and ങ്ങ (ñña), the numbers of loops are 0, 1, 2 and 3 respectively. Different types of loops appear in handwritten patterns and are displayed in Fig. 2. This feature is calculated from the character skeleton.

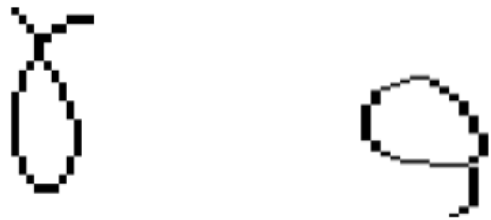

**Fig. 2.** Samples of loops in handwritten pattern.

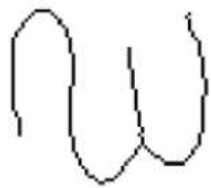

**Fig. 3.** Character skeleton ഡ (/da/)

### 3.1.2 Number of endpoints:

An end point is defined as the start or end of a line segment. A foreground pixel is considered to be an endpoint if it has exactly one foreground neighbour in the 3 × 3 neighbourhood. For example, in the case of character ப (/pa/), the number of end points is 2. The number of end points in the skeleton of character pattern ட(/ḍa/) of Fig. 3 is 3. This feature is extracted from the character skeleton.

### 3.1.3 Number of branch points:

A branch point is a junction point that joins three branches. A foreground pixel is considered to be a branch point if it has exactly three foreground neighbours in the 3 × 3 neighbourhood. For example, there is only one branch point for character image ധ (/dha/). The number of branch points in the skeleton of character pattern ட(/ḍa/ ) of Fig. 3 is 1. This feature is computed from character skeleton.

### 3.1.4 Number of cross points:

A cross point is a junction point connecting four branches. A foreground pixel is considered to be a cross point if it has exactly four foreground neighbors in the 3 × 3 neighbourhood. For example, there is exactly one cross point in ழ (/zha/). The number of cross points in the skeleton of character pattern ட(/ḍa/) of Fig. 3 is 0. This feature is identified from the skeleton representation of the character pattern.

### 3.1.5 Ratio of width to height:

The width/height ratio is calculated based on the bounding box, which is the rectangle enclosing the character pattern. Some of the characters have width greater than height, some have width less than height and some others have approximately equal width and height. This feature is calculated from the original segmented image.

## 3.2 Distribution of Foreground Pixels

For a character image, depending on the shape of the pattern, the pixels are distributed unevenly within the normalized window. To identify characters the pattern image is divided into a number of equal sized or variable sized regions or zones. This technique is referred to as zoning. From each such region, features are extracted. Fig. 4 demonstrates the division of character pattern ಠ(/ra/) into 4×4 equal zones

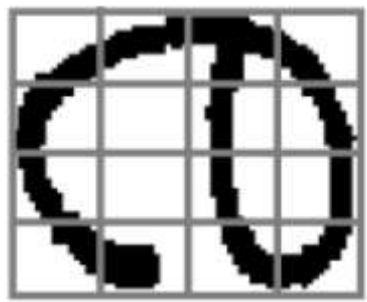

**Fig. 4.** Character pattern ಠ(/ra/) divided into 4 × 4 zones

For creating distribution features, each handwritten character pattern is divided into a set of zones using a uniform grid of arbitrary size, and from each zone, the sum of foreground (black) pixels is computed. These features provide an idea about the local distribution of pixels in each zone. The number of local features can be arbitrarily determined by changing the number of zones by varying the size of the grid. The optimal number of zone sizes is determined experimentally.

## 3.3 Transition Count Features

Transition count is defined as the number of transitions from a foreground pixel to background pixel along vertical and horizontal lines through a character pattern. It is computed directly from the binary image. Let the character image be scanned along each horizontal scan line from top to bottom. For each scan line or row, we count the number of transitions from foreground pixel to background pixel. Let the horizontal transition count, for the i<sup>th</sup> scan line, be **T<sub>h</sub>(i)**. If N is the total number of scan lines, the sequence **{T<sub>h</sub>(i); i=1,2,…N}** can be treated as the horizontal transition count vector of the character image. To normalize the transition count, we divide each **T<sub>h</sub>(i)** by N, the number of scan lines, i.e. **NT<sub>h</sub>(i)=(T<sub>h</sub>(i)/N).** Normalized horizontal transition count vector of the character image is defined as: **NT<sub>H</sub>( I )={ NT<sub>h</sub>(i); i=1,2,… N}.** Similarly, a normalized vertical transition count vector **NT<sub>V</sub>( I )** is also created. The transition count feature descriptor for the character image is defined as: TC(I)={ **NT<sub>H</sub>( I ), NTv( I )}**

### 3.4 Local Binary Pattern Descriptors

Local Binary Pattern Descriptor (LBP) is a texture descriptor. The LBP operator is denoted as LBP$_{P,R}$, where R is the radius of the neighbourhoods and P is the number of neighbourhoods. To calculate LBP for each pixel, its value is compared to each of its eight neighours. The comparison gives either a 1 or a 0 indicating whether the centre pixel's value is greater than its neighbours. To create feature vectors, we find out the LBP code for each pixel in the image. Using this LBP image, we compute the horizontal and vertical projection profiles.

### 4 Classification

For classification k-NN, SVM and ELM have been employed to see which one performs best. The feature values are min-max normalized in the range -1, +1. KNN is a non parametric lazy learning algorithm which makes a decision based on the entire training data. Basically this classifier finds k nearest neighbours from the entire training data and does a majority voting for a decision. SVM is one of the popular techniques for pattern recognition for linear and non-linear classification [8]. SVM can achieve high generalization accuracies when trained with a large number of samples. We have used SVM-RBF and Polynomial in this work. The ELM [9] has recently been proposed for a single-hidden layer feedforward neural network that achieves good generalization performance at extremely fast learning speed with random input weights. It aims to reach the smallest training error and also the smallest norm of output weights.

## 5 Results and Discussions

### 5.1 Malayalam handwritten database and Experimental Setup

For experimentation, a bench-mark database containing all the 44 basic characters; 5 chillu (pure consonants), 26 compound characters and 15 vowel-consonant signs were used. Altogether, 90 classes were taken. For uniform distribution of samples per class, 150 samples per class are randomly chosen to create the training database and from the remaining set, 50 samples are randomly chosen to create the testing set. Training set contains 13500 samples and test set contains 4500 samples and the size of the whole dataset is 18000. Random permutations of samples are performed both on training set and testing set. The training set and testing set are disjoint and the same training set and testing set are used throughout the experiments.

### 5.2 Result and Analysis

The classification result using individual features is depicted in Table 1. As the features chosen are complementary in nature we have combined them. As the dimension of the feature vector becomes 325, using Principal Component Analysis, we have reduced the number of features to 76, beyond that contribution becomes negligible. Scree plot showing the variance explained by principal components is depicted in Fig. 5. Classification results in PCA feature space are provided in Table 2. The recognition accuracy of all the classifiers has been increased using the combined features. The highest accuracy of **92.31%** is obtained with the SVM-RBF kernel with a testing time of 27.59 seconds.

Table 1. Recognition Accuracy on 90 character classes (%)

| Description of Features | Dim. | SVM-RBF | SVM-Poly | ELM-noReg | ELM-opt | k-NN |
|---|---|---|---|---|---|---|
| Topological Features | 64 | 84.91 | 86.15 | 79.67 | 81.42 | 75.02 |
| Topological Features and Distribution Features | 69 | 86.47 | 87.44 | 81.42 | 82.36 | 76.11 |
| Transition Count Features | 128 | 76.73 | 74.6 | 67.64 | 67.60 | 67.64 |
| LBP Features | 128 | 58.35 | 51.75 | 48.49 | 51.00 | 48.78 |

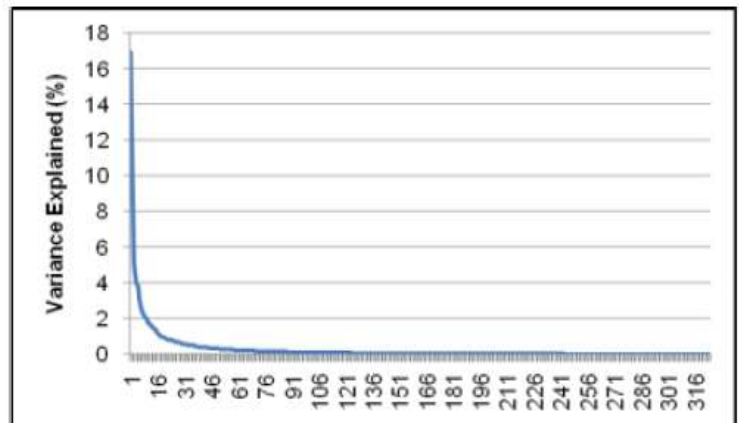

**Fig. 5.** Scree Plot: Variance Explained by Principal Components

**Table 2.** Classification Result of Feature Combination in PCA Feature Space

| Feature Combination<br>Reduced Dimension : 76 | SVM-RBF | SVM-Poly | k-NN<br>k=5 | ELM-<br>noReg | ELM-<br>opt |
|---|---|---|---|---|---|
| Training Accuracy (%) | 99.97 | 99.96 | - | 98.16 | 97.88 |
| Testing Accuracy (%) | 92.31 | 91.53 | 84.78 | 87.27 | 89.0 |
| Training Time in seconds | 87.53 | 66.59 | - | 536 | 262.97 |
| Testing Time in seconds | 27.59 | 17.77 | 68.28 | 3.5 | 5.72 |

## 6. Conclusion

In this paper, we have represented a set of spatial domain features suitable for recognition. The features selected include topological features, distribution features, transition count features, LBP features. Topological features provide global aspects of a character while distribution features take care of statistical distribution of pixels in local regions. Transition count features provide the information such as switching from foreground pixels to background in each horizontal and vertical scan line. As these informations are complementary to one another, we have decided to combine the features. The combined features performed very well in the recognition stage. The recognition accuracy of 92.31% was obtained with the SVM-RBF classifier with a testing time of 27.59 seconds.